\documentclass[12pt,a4paper]{article}
\usepackage[utf8]{inputenc}
\usepackage[english]{babel}
\usepackage{mathptmx} %
\usepackage{color} 
\usepackage{xcolor,colortbl}
\usepackage[center]{caption}
\usepackage{tcolorbox}
\usepackage{tabularx}
\usepackage{makecell}

\definecolor{blur.corr}{RGB}{50, 110, 30}
\definecolor{noise.corr}{RGB}{145, 105, 70}
\definecolor{weather.corr}{RGB}{80, 170, 200}
\definecolor{digital.corr}{RGB}{175, 110, 150}

\usepackage{wrapfig}
\usepackage{amsmath}
\usepackage{amsfonts}
\usepackage{amssymb}
\usepackage{graphicx}
\usepackage{subfig}
\usepackage{enumitem}
\usepackage{changepage}
\usepackage[colorinlistoftodos]{todonotes}
\usepackage[colorlinks=true,citecolor=black,linkcolor=black,urlcolor=black]{hyperref}
\usepackage{authblk}
\usepackage{float}
\usepackage{todonotes}
\usepackage{fullpage}
\usepackage[font={small,it}]{caption}
\usepackage{lineno}
\usepackage[margin=10pt,font=small,labelfont=bf]{caption}
\usepackage{multirow,booktabs,setspace,caption}
\usepackage{scalerel,stackengine}
\usepackage{rotating}
\usepackage{epstopdf}

\DeclareCaptionLabelSeparator*{spaced}{\\[2ex]}
\newcommand{\paragraphwithlinebreak}[1]{\paragraph{#1}\mbox{}\\}

\newcommand\blfootnote[1]{%
	\begingroup
	\renewcommand\thefootnote{}\footnote{#1}%
	\addtocounter{footnote}{-1}%
	\endgroup
}

\graphicspath{{figures/}}

\author[]{
Robert Geirhos\textsuperscript{1,2,$\ast$,$\S$},
Jörn-Henrik Jacobsen\textsuperscript{3,$\ast$},
Claudio Michaelis\textsuperscript{1,2,$\ast$},\\
Richard Zemel\textsuperscript{$\dagger$,3},
Wieland Brendel\textsuperscript{$\dagger$,1},
Matthias Bethge\textsuperscript{$\dagger$,1} \&
Felix A. Wichmann\textsuperscript{$\dagger$,}}

\affil[1]{University of Tübingen, Germany}
\affil[2]{International Max Planck Research School for Intelligent Systems, Germany}
\affil[3]{University of Toronto, Vector Institute, Canada}
\affil[$\ast$]{Joint first / $\dagger$ joint senior authors}
\affil[$\S$]{To whom correspondence should be addressed: \texttt{robert.geirhos@wichmannlab.org}}

\title{Shortcut Learning in Deep Neural Networks}

\begin{document}
\date{}

\maketitle

\abstract{
\noindent
Deep learning has triggered the current rise of artificial intelligence and is the workhorse of today's machine intelligence. Numerous success stories have rapidly spread all over science, industry and society, but its limitations have only recently come into focus. In this perspective we seek to distill how many of deep learning's problems can be seen as different symptoms of the same underlying problem: \emph{shortcut learning}. Shortcuts are decision rules that perform well on standard benchmarks but fail to transfer to more challenging testing conditions, such as real-world scenarios. Related issues are known in Comparative Psychology, Education and Linguistics, suggesting that shortcut learning may be a common characteristic of learning systems, biological and artificial alike. Based on these observations, we develop a set of recommendations for model interpretation and benchmarking, highlighting recent advances in machine learning to improve robustness and transferability from the lab to real-world applications.
}

\section{Introduction}
If science was a journey, then its destination would be the discovery of simple explanations to complex phenomena. There was a time when the existence of tides, the planet's orbit around the sun, and the observation that ``things fall down'' were all largely considered to be independent phenomena---until 1687, when Isaac Newton formulated his law of gravitation that provided an elegantly simple explanation to all of these (and many more). Physics has made tremendous progress over the last few centuries, but the thriving field of deep learning is still very much at the beginning of its journey---often lacking a detailed understanding of the underlying principles.\blfootnote{This is the preprint version of an article that has been published by Nature Machine Intelligence (\href{https://doi.org/10.1038/s42256-020-00257-z}{https://doi.org/10.1038/s42256-020-00257-z}).}

For some time, the tremendous success of deep learning has perhaps overshadowed the need to thoroughly understand the behaviour of Deep Neural Networks (DNNs). In an ever-increasing pace, DNNs were reported as having achieved human-level object classification performance \cite{he2015delving}, beating world-class human Go, Poker, and Starcraft players \cite{silver2016mastering, moravvcik2017deepstack}, detecting cancer from X-ray scans \cite{rajpurkar2017chexnet}, translating text across languages \cite{devlin2018bert}, helping combat climate change \cite{rolnick2019tackling}, and accelerating the pace of scientific progress itself \cite{reichstein2019deep}. Because of these successes, deep learning has gained a strong influence on our lives and society. At the same time, however, researchers are unsatisfied about the lack of a deeper understanding of the underlying principles and limitations. Different from the past, tackling this lack of understanding is not a purely scientific endeavour anymore but has become an urgent necessity due to the growing societal impact of machine learning applications. If we are to trust algorithms with our lives by being driven in an autonomous vehicle, if our job applications are to be evaluated by neural networks, if our cancer screening results are to be assessed with the help of deep learning---then we indeed need to understand thoroughly: When does deep learning work? When does it fail, and why?

\begin{figure*}[t]
    \begin{center}
    \includegraphics[width=0.95\linewidth]{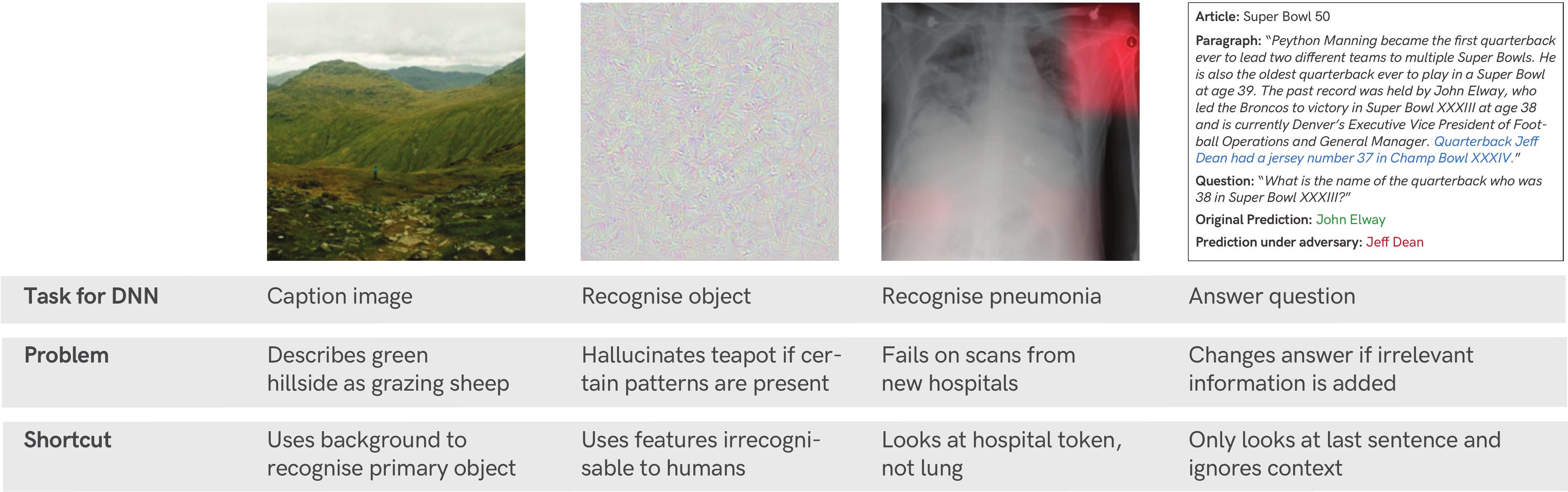}
    \end{center}
    \caption{Deep neural networks often solve problems by taking shortcuts instead of learning the intended solution, leading to a lack of generalisation and unintuitive failures. This pattern can be observed in many real-world applications.}
\label{fig:teaser}
\end{figure*}

In terms of understanding the limitations of deep learning, we are currently observing a large number of failure cases, some of which are visualised in Figure~\ref{fig:teaser}. DNNs achieve super-human performance recognising objects, but even small invisible changes \cite{szegedy2013intriguing} or a different background context \cite{beery2018recognition, rosenfeld2018elephant} can completely derail predictions. DNNs can generate a plausible caption for an image, but---worryingly---they can do so without ever looking at that image \cite{heuer2016generating}.
DNNs can accurately recognise faces, but they show high error rates for faces from minority groups \cite{buolamwini2018gender}. DNNs can predict hiring decisions on the basis of résumés, but the algorithm's decisions are biased towards selecting men \cite{dastin2018amazon}.

How can this discrepancy between super-human performance on one hand and astonishing failures on the other hand be reconciled? One central observation is that many failure cases are not independent phenomena, but are instead connected in the sense that DNNs follow unintended ``shortcut'' strategies. While superficially successful, these strategies typically fail under slightly different circumstances. For instance, a DNN may appear to classify cows perfectly well---but fails when tested on pictures where cows appear outside the typical grass landscape, revealing ``grass'' as an unintended (shortcut) predictor for ``cow'' \cite{beery2018recognition}. Likewise, a language model may appear to have learned to reason---but drops to chance performance when superficial correlations are removed from the dataset \cite{niven2019probing}. Worse yet, a machine classifier successfully detected pneumonia from X-ray scans of a number of hospitals, but its performance was surprisingly low for scans from novel hospitals: The model had unexpectedly learned to identify particular hospital systems with near-perfect accuracy (e.g.\ by detecting a hospital-specific metal token on the scan, see Figure~\ref{fig:teaser}). Together with the hospital's pneumonia prevalence rate it was able to achieve a reasonably good prediction---without learning much about pneumonia at all \cite{zech2018variable}.

At a principal level, shortcut learning is not a novel phenomenon. The field of machine learning with its strong mathematical underpinnings has long aspired to develop a formal understanding of shortcut learning which has led to a variety of mathematical concepts and an increasing amount of work under different terms such as \emph{learning under covariate shift} \cite{bickel2009discriminative}, \emph{anti-causal learning} \cite{scholkopf2012causal}, \emph{dataset bias} \cite{torralba2011unbiased}, the \emph{tank legend} \cite{branwen2011the} and the \emph{Clever Hans effect} \cite{pfungst1911clever}. This perspective aims to present a unifying view of the various phenomena that can be collectively termed shortcuts, to describe common themes underlying them, and lay out the approaches that are being taken to address them both in theory and in practice.

The structure of this perspective is as follows. Starting from an intuitive level, we introduce shortcut learning across biological neural networks (Section~\ref{sec:biological_shortcuts}) and then approach a more systematic level by introducing a taxonomy (Section~\ref{sec:taxonomy}) and by investigating the origins of shortcuts (Section~\ref{sec:characteristics}). In Section~\ref{sec:across_deep_learning}, we highlight how these characteristics affect different areas of deep learning (Computer Vision, Natural Language Processing, Agent-based Learning, Fairness). The remainder of this perspective identifies actionable strategies towards diagnosing and understanding shortcut learning (Section~\ref{sec:diagnosing_understanding}) as well as current research directions attempting to overcome shortcut learning (Section~\ref{sec:beyond_shortcut_learning}). Overall, our selection of examples is biased towards Computer Vision since this is one of the areas where deep learning has had its biggest successes, and an area where examples are particularly easy to visualise. We hope that this perspective facilitates the awareness for shortcut learning and motivates new research to tackle this fundamental challenge we currently face in machine learning.

\section{Shortcut learning in biological neural networks}
\label{sec:biological_shortcuts}
Shortcut learning typically reveals itself by a strong discrepancy between intended and actual learning strategy, causing an unexpected failure. Interestingly, machine learning is not alone with this issue: From the way students learn to the unintended strategies rats use in behavioural experiments---variants of shortcut learning are also common for biological neural networks. We here point out two examples of unintended learning strategies by natural systems in the hope that this may provide an interesting frame of reference for thinking about shortcut learning within and beyond artificial systems.

\subsection{Shortcut learning in Comparative Psychology: unintended cue learning}
\label{subsec:comparative_psychology}

\emph{Rats learned to navigate a complex maze apparently based on subtle colour differences---very surprising given that the rat retina has only rudimentary machinery to support at best somewhat crude colour vision. Intensive investigation into this curious finding revealed that the rats had tricked the researchers: They did not use their visual system at all in the experiment and instead simply discriminated the colours by the odour of the colour paint used on the walls of the maze. Once smell was controlled for, the remarkable colour discrimination ability disappeared ...}\footnote{Nicholas Rawlins, personal communication with F.A.W. some time in the early 1990s, confirmed via email on 12.11.2019.}\\

\noindent
Animals are no strangers to finding simple, unintended solutions that fail unexpectedly: They are prone to \emph{unintended cue learning}, as shortcut learning is called in Comparative Psychology and the Behavioural Neurosciences. When discovering cases of unintended cue learning, one typically has to acknowledge that there was a crucial difference between performance in a given experimental paradigm (e.g.\ rewarding rats to identify different colours) and the investigated mental ability one is actually interested in (e.g.\ visual colour discrimination). In analogy to machine learning, we have a striking discrepancy between intended and actual learning outcome.

\subsection{Shortcut learning in Education: surface learning}
\label{subsec:education}
\emph{Alice loves history. Always has, probably always will. At this very moment, however, she is cursing the subject: After spending weeks immersing herself in the world of Hannibal and his exploits in the Roman Empire, she is now faced with a number of exam questions that are (in her opinion) to equal parts dull and difficult. ``How many elephants did Hannibal employ in his army---19, 34 or 40?'' ... Alice notices that Bob, sitting in front of her, seems to be doing very well. Bob of all people, who had just boasted how he had learned the whole book chapter by rote last night ...}\\

\noindent
In educational research, Bob's reproductive learning strategy would be considered \emph{surface learning}, an approach that relies on narrow testing conditions where simple discriminative generalisation strategies can be highly successful. This fulfils the characteristics of shortcut learning by giving the appearance of good performance but failing immediately under more general test settings. Worryingly, surface learning helps rather than hurts test performance on typical multiple-choice exams \cite{scouller1998influence}: Bob is likely to receive a good grade, and judging from grades alone Bob would appear to be a much better student than Alice in spite of her focus on understanding. Thus, in analogy to machine learning we again have a striking discrepancy between intended and actual learning outcome.

\section{Shortcuts defined: a taxonomy of decision rules}
\label{sec:taxonomy}

\begin{figure*}[t]
    \begin{center}
    \includegraphics[width=0.95\linewidth]{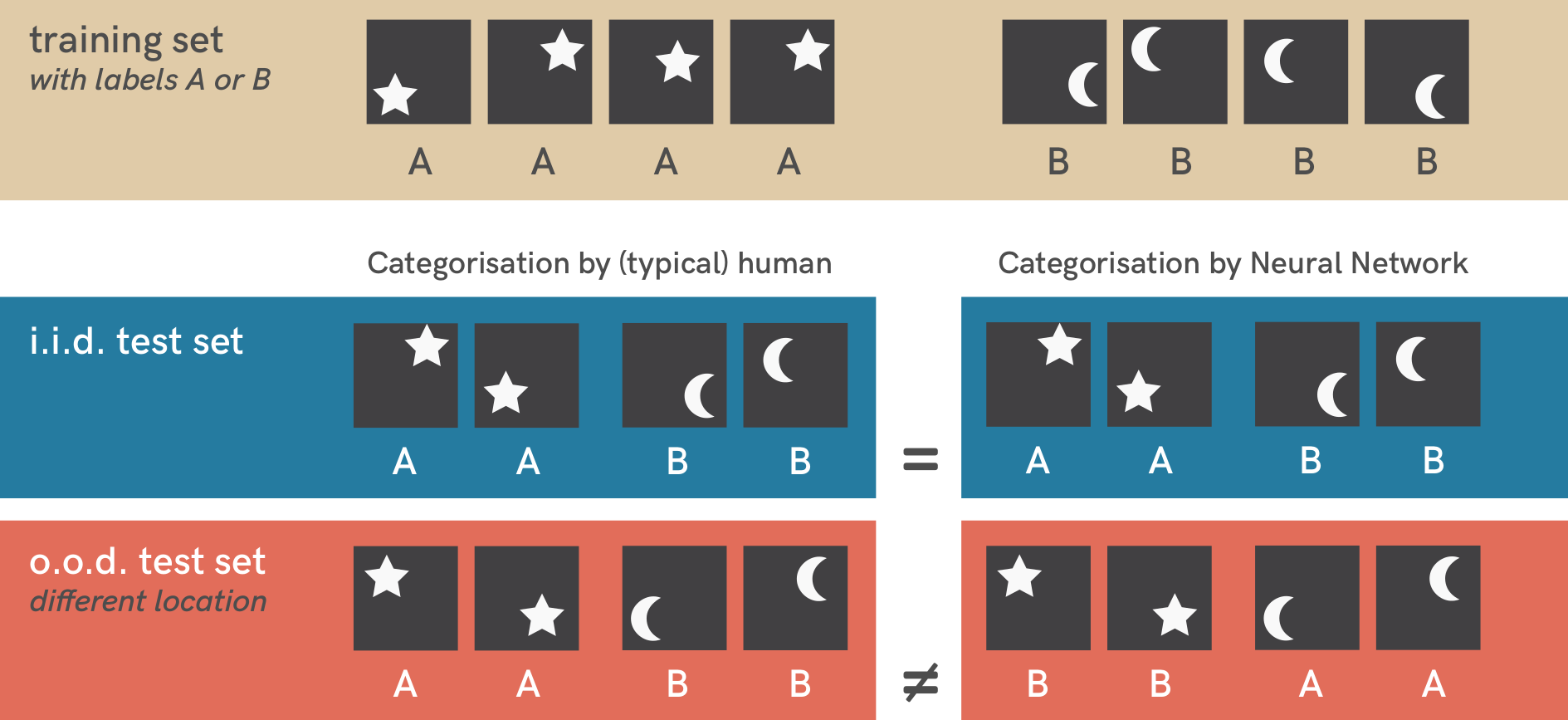}
    \end{center}
    \caption{Toy example of shortcut learning in neural networks. When trained on a simple dataset of stars and moons (top row), a standard neural network (three layers, fully connected) can easily categorise novel similar exemplars (mathematically termed i.i.d.\ test set, defined later in Section~\ref{sec:taxonomy}). However, testing it on a slightly different dataset (o.o.d.\ test set, bottom row) reveals a shortcut strategy: The network has learned to associate object location with a category. During training, stars were always shown in the top right or bottom left of an image; moons in the top left or bottom right. This pattern is still present in samples from the i.i.d.\ test set (middle row) but not in o.o.d.\ test images (bottom row), exposing the shortcut.}
\label{fig:toy_example}
\end{figure*}

With examples of biological shortcut learning in mind (examples which we will return to in Section~\ref{sec:diagnosing_understanding}), what does shortcut learning in artificial neural networks look like? Figure~\ref{fig:toy_example} shows a simple classification problem that a neural network is trained on (distinguishing a star from a moon).\footnote{Code is available from \url{https://github.com/rgeirhos/shortcut-perspective}.} When testing the model on similar data (blue) the network does very well---or so it may seem. Very much like the smart rats that tricked the experimenter, the network uses a shortcut to solve the classification problem by relying on the location of stars and moons. When location is controlled for, network performance deteriorates to random guessing (red). In this case (as is typical for object recognition), classification based on object shape would have been the intended solution, even though the difference between intended and shortcut solution is not something a neural network can possibly infer from the training data. 

\begin{figure*}[t]
    \begin{center}
    \includegraphics[width=\linewidth]{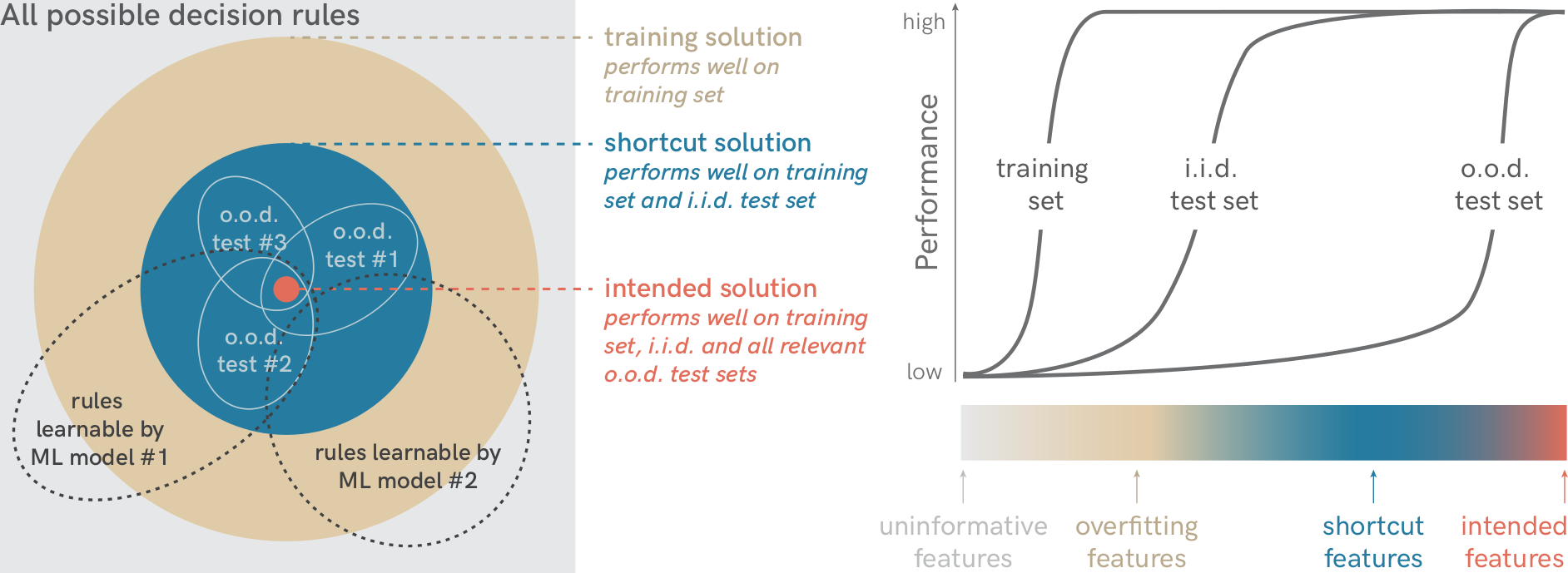}
    \end{center}
    \caption{Taxonomy of decision rules. Among the set of all possible rules, only some solve the training data. Among the solutions that solve the training data, only some generalise to an i.i.d.\ test set. Among those solutions, shortcuts fail to generalise to different data (o.o.d.\ test sets), but the intended solution does generalise.}
\label{fig:taxonomy}
\end{figure*}

On a general level, any neural network (or machine learning algorithm) implements a decision rule which defines a relationship between input and output---in this example assigning a category to every input image. Shortcuts, the focus of this article, are one particular group of decision rules. In order to distinguish them from other decision rules, we here introduce a taxonomy of decision rules (visualised in Figure~\ref{fig:taxonomy}), starting from a very general rule and subsequently adding more constraints until we approach the intended solution.

\paragraphwithlinebreak{(1) all possible decision rules, including non-solutions}
Imagine a model that tries to solve the problem of separating stars and moons by predicting ``star'' every time it detects a white pixel in the image. This model uses an \emph{uninformative feature} (the grey area in Figure~\ref{fig:taxonomy}) and does not reach good performance on the data it was trained on, since it implements a poor decision rule (both moon and star images contain white pixels). Typically, interesting problems have an abundant amount of non-solutions.

\paragraphwithlinebreak{(2) training solutions, including overfitting solutions}
In machine learning it is common practice to split the available data randomly into a training and a test set. The training set is used to guide the model in its selection of a (hopefully useful) decision rule, and the test set is used to check whether the model achieves good performance on similar data it has not seen before. Mathematically, the notion of similarity between training and test set  commonly referred to in machine learning is the assumption that the samples in both sets are drawn from the same distribution. This is the case if both the data generation mechanism and the sampling mechanism are identical. In practice this is achieved by randomising the split between training and test set. The test set is then called independent and identically distributed (i.i.d.) with regard to the training set. In order to achieve high average performance on the test set, a model needs to learn a function that is approximately correct within a subset of the input domain which covers most of the probability of the distribution. If a function is learned that yields the correct output on the training images but not on the i.i.d. test images, the learning machine uses \emph{overfitting features} (the blue area in Figure~\ref{fig:taxonomy}). 

\paragraphwithlinebreak{(3) i.i.d.\ test solutions, including shortcuts}
Decision rules that solve both the training and i.i.d.\ test set typically score high on standard benchmark leaderboards. However, even the simple toy example can be solved through at least three different decision rules: (a) by shape, (b) by counting the number of white pixels (moons are smaller than stars) or (c) by location (which was correlated with object category in the training and i.i.d.\ test sets). As long as tests are performed only on i.i.d.\ data, it is impossible to distinguish between these. However, one can instead test models on datasets that are systematically different from the i.i.d.\ training and test data (also called \emph{out-of-distribution} or \emph{o.o.d.} data). For example, an o.o.d.\ test set with randomised object size will instantly invalidate a rule that counts white pixels. Which decision rule is the \emph{intended solution} is clearly in the eye of the beholder, but humans often have clear expectations. In our toy example, humans typically classify by shape. A standard fully connected neural network\footnote{A convolutional (rather than fully connected) network would be prevented from taking this shortcut by design.} trained on this dataset, however, learns a location-based rule (see Figure~\ref{fig:toy_example}). In this case, the network has used a \emph{shortcut feature} (the blue area in Figure~\ref{fig:taxonomy}): a feature that helps to perform well on i.i.d.\ test data but fails in o.o.d.\ generalisation tests.

\paragraphwithlinebreak{(4) intended solution}
Decision rules that use the \emph{intended features} (the red area in Figure~\ref{fig:taxonomy}) work well not only on an i.i.d.\ test set but also perform as intended on o.o.d.\ tests, where shortcut solutions fail. In the toy example, a decision rule based on object shape (the intended feature) would generalise to objects at a different location or with a different size. Humans typically have a strong intuition for what the intended solution should be capable of. Yet, for complex problems, intended solutions are mostly impossible to formalise, so machine learning is needed to estimate these solutions from examples. Therefore the choice of examples, among other aspects, influence how closely the intended solution can be approximated.\\

\section{Shortcuts: where do they come from?}
\label{sec:characteristics}
Following this taxonomy, shortcuts are decision rules that perform well on i.i.d.\ test data but fail on o.o.d.\ tests, revealing a mismatch between intended and learned solution. It is clear that shortcut learning is to be avoided, but where do shortcuts come from, and what are the defining real-world characteristics of shortcuts that one needs to look out for when assessing a model or task through the lens of shortcut learning? There are two different aspects that one needs to take into account. First, shortcut opportunities (or shortcut features) in the data: possibilities for solving a problem differently than intended (Section~\ref{subsec:dataset_shortcut_opportunities}). Second, feature combination: how different features are combined to form a decision rule (Section~\ref{subsec:feature_combination}). Together, these aspects determine how a model generalises (Section~\ref{subsec:generalisation_robustness}).

\subsection{Dataset: shortcut opportunities}
\label{subsec:dataset_shortcut_opportunities}

\newcolumntype{C}[1]{>{\centering\let\newline\\\arraybackslash\hspace{0pt}}m{#1}}
\newcolumntype{L}[1]{>{\let\newline\\\arraybackslash\hspace{0pt}}m{#1}}
\begin{tabular}{C{2.8cm}  L{10.4cm}}
    \frame{\includegraphics[width=\linewidth]{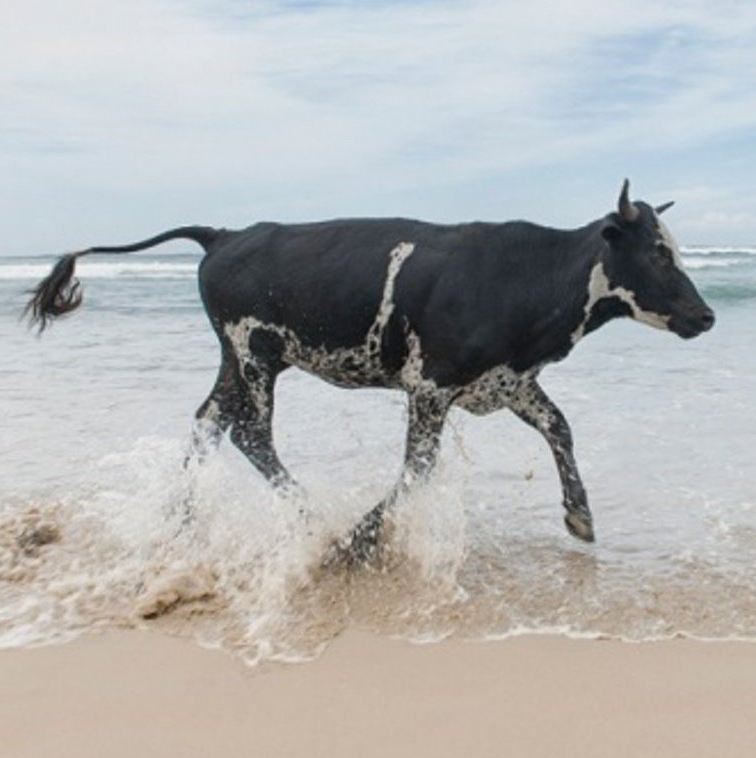}} & What makes a cow a cow? To DNNs, a familiar background can be as important for recognition as the object itself, and sometimes even more important: A cow at an unexpected location (such as a beach rather than grassland) is not classified correctly \cite{beery2018recognition}. Conversely, a lush hilly landscape without any animal at all might be labelled as a ``herd of grazing sheep'' by a DNN \cite{shane2018do}.\\
\end{tabular}

\noindent
This example highlights how a systematic relationship between object and background or context can easily create a shortcut opportunity. If cows happen to be on grassland for most of the training data, detecting grass instead of cows becomes a successful strategy for solving a classification problem in an unintended way; and indeed many models base their predictions on context \cite{wichmann2010animal, ribeiro2016should, zhu2016object, wang2017visual, beery2018recognition, dawson2018same, rosenfeld2018elephant}. Many shortcut opportunities are a consequence of natural relationships, since grazing cows are typically surrounded by grassland rather than water. These so-called \emph{dataset biases} have long been known to be problematic for machine learning algorithms \cite{torralba2011unbiased}. Humans, too, are influenced by contextual biases (as evident from faster reaction times when objects appear in the expected context), but their predictions are much less affected when context is missing \cite{biederman1981semantics, biederman1982scene, oliva2007role, castelhano2011scene}. In addition to shortcut opportunities that are fairly easy to recognise, deep learning has led to the discovery of much more subtle shortcut features, including high-frequency patterns that are almost invisible to the human eye \cite{jo2017measuring, ilyas2019adversarial}. Whether easy to recognise or hard to detect, it is becoming more and more evident that shortcut opportunities are by no means disappearing when the size of a dataset is simply scaled up by some orders of magnitude (in the hope that this is sufficient to sample the diverse world that we live in \cite{halevy2009unreasonable}). Systematic biases are still present even in ``Big Data'' with large volume and variety, and consequently even large real-world datasets usually contain numerous shortcut opportunities. Overall, it is quite clear that data alone rarely constrains a model sufficiently, and that data cannot replace making assumptions \cite{wolpert1997no}. The totality of all assumptions that a model incorporates (such as, e.g., the choice of architecture) is called the \emph{inductive bias} of a model and will be discussed in more detail in Section~\ref{subsec:shortcuts_why_learned}.

\subsection{Decision rule: shortcuts from discriminative learning}
\label{subsec:feature_combination}

\begin{tabular}{C{2.8cm}  L{10.4cm}}
    \frame{\includegraphics[width=\linewidth]{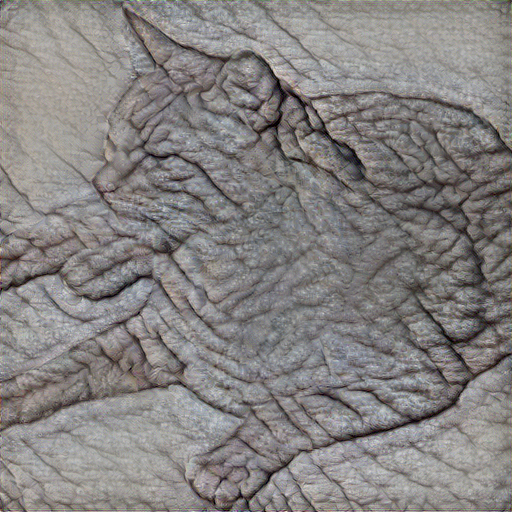}} & What makes a cat a cat? To standard DNNs, the example image on the left clearly shows an elephant, not a cat. Object textures and other local structures in images are highly useful for object classification in standard datasets \cite{Brendel2019approximating}, and DNNs strongly rely on texture cues for object classification, largely ignoring global object shape \cite{baker2018deep, geirhos2019imagenettrained}.\\
\end{tabular}

\noindent
In many cases, relying on object textures can be sufficient to solve an object categorisation task. Obviously, however, texture is only one of many attributes that define an object. Discriminative learning differs from generative modeling by picking any feature that is sufficient to reliably discriminate on a given dataset but the learning machine has no notion of how realistic examples typically look like and how the features used for discrimination are combined with other features that define an object. In our example, using textures for object classification becomes problematic if other intended attributes (like shape) are ignored entirely. This exemplifies the importance of feature combination: the definition of an object relies on a (potentially highly non-linear) combination of information from different sources or attributes that influence a decision rule.\footnote{In Cognitive Science, this process is called \emph{cue combination}.} In the example of the cat with elephant texture above, a shape-agnostic decision rule that merely relies on texture properties clearly fails to capture the task of object recognition as it is understood for human vision. While the model uses an important attribute (texture) it tends to equate it with the definition of the object missing out other important attributes such as shape. Of course, being aligned with the human decision rule does not always conform to our intention. In medical or safety-critical applications, for instance, we may instead seek an improvement over human performance.

Inferring human-interpretable object attributes like shape or texture from an image requires specific nonlinear computations. In typical end-to-end discriminative learning, this again may be prone to shortcut learning. Standard DNNs do not impose any human-interpretability requirements on intermediate image representations and thus might be severely biased to the extraction of overly simplistic features which only generalise under the specific design of the particular dataset used but easily fail otherwise. Discriminative feature learning goes as far that some decision rules only depend on a single predictive pixel \cite{heinze2017conditional, malhotra2018difference, jacobsen2019excessive} while all other evidence is ignored.\footnote{In models of animal learning, the \emph{blocking effect} is a related phenomenon. Once a predictive cue/feature (say, a light flash) has been associated with an outcome (e.g.\ food), animals sometimes fail to associate a new, equally predictive cues with the same outcome \cite{kamin1969predictability, dickinson1980contemporary, bouton2007learning}.} In principle, ignoring some evidence can be beneficial. In object recognition, for example, we want the decision rule  to be invariant to an object shift. However, undesirable invariance (sometimes called \emph{excessive invariance}) is harmful and modern machine learning models can be invariant to almost all features that humans would rely on when classifying an image \cite{jacobsen2019excessive}.

\subsection{Generalisation: how shortcuts can be revealed}
\label{subsec:generalisation_robustness}

\begin{tabular}{C{2.8cm}  L{10.4cm}}
    \frame{\includegraphics[width=\linewidth]{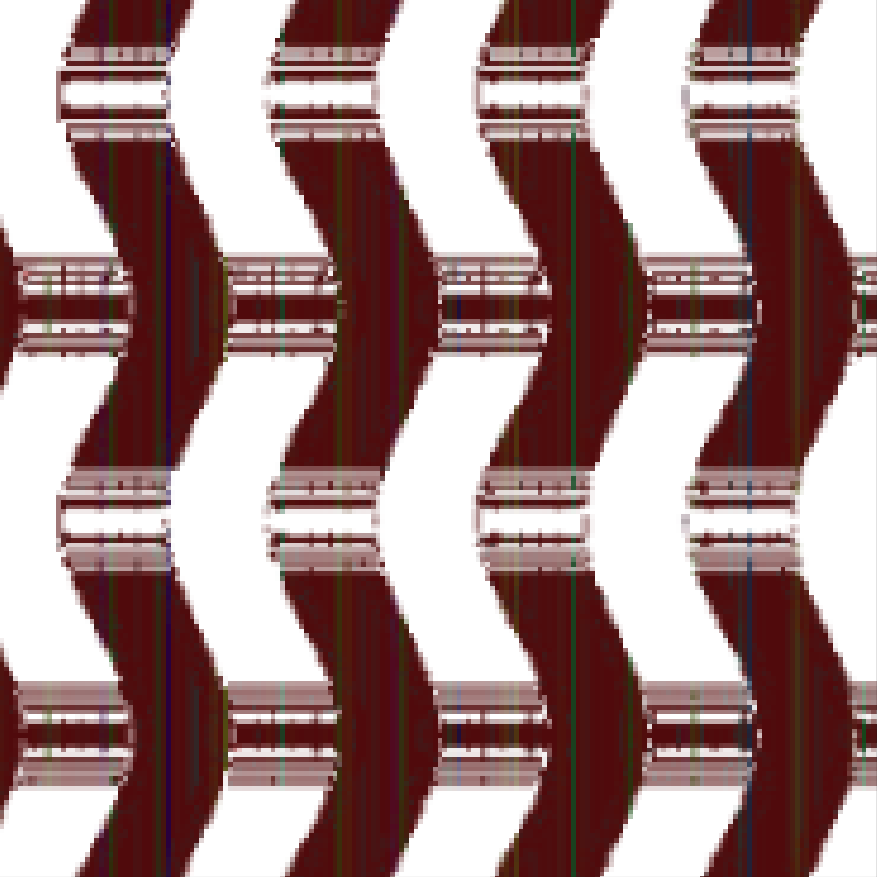}} & What makes a guitar a guitar? When tested on this pattern never seen before, standard DNNs predict ``guitar'' with high certainty \cite{nguyen2015deep}. Exposed by the generalisation test, it seems that DNNs learned to detect certain patterns (curved guitar body? strings?) instead of guitars: a successful strategy on training and i.i.d.\ test data that leads to unintended generalisation on o.o.d.\ data.
\end{tabular}

\begin{figure}[t]
	\includegraphics[width=1\textwidth]{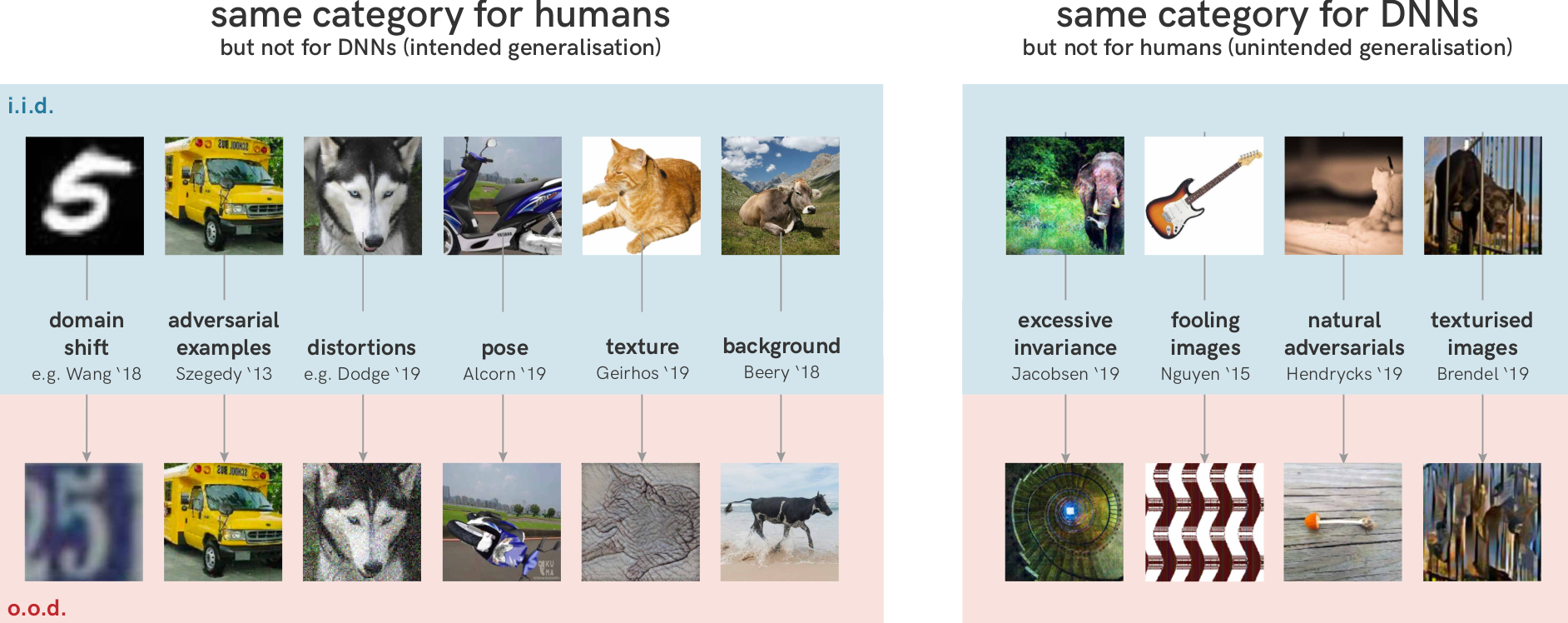}
	\caption{Both human and machine vision generalise, but they generalise very differently. Left: image pairs that belong to the same category for humans, but not for DNNs. Right: images pairs assigned to the same category by a variety of DNNs, but not by humans.}
	\label{fig:generalisation}
\end{figure}

\noindent
This exemplifies the inherent link between shortcut learning and generalisation. By itself, generalisation is not a part of shortcut learning---but more often than not, shortcut learning is discovered through cases of unintended generalisation, revealing a mismatch between human-intended and model-learned solution. Interestingly, DNNs do not suffer from a general lack of o.o.d.\ generalisation (Figure~\ref{fig:generalisation}) \cite{nguyen2015deep, Brendel2019approximating, hendrycks2019natural, jacobsen2019excessive}. DNNs recognise guitars even if only some abstract pattern is left---however, this remarkable generalisation performance is undesired, at least in this case. In fact, the set of images that DNNs classify as ``guitar'' with high certainty is incredibly big. To humans only some of these look like guitars, others like patterns (interpretable or abstract) and many more resemble white noise or even look like airplanes, cats or food \cite{szegedy2013intriguing, nguyen2015deep, jacobsen2019excessive}. Figure~\ref{fig:generalisation} on the right, for example, highlights a variety of image pairs that have hardly anything in common for humans but belong to the same category for DNNs. Conversely, to the human eye an image's category is not altered by innocuous \emph{distribution shifts} like rotating objects or adding a bit of noise, but if these changes interact with the shortcut features that DNNs are sensitive to, they completely derail neural network predictions \cite{szegedy2013intriguing, azulay2018deep, beery2018recognition, wang2018deep, alcorn2019strike, dodge2019human, geirhos2019imagenettrained}. This highlights that generalisation failures are neither a failure to learn nor a failure to generalise at all, but instead a failure to generalise in the intended direction---generalisation and robustness can be considered the flip side of shortcut learning. Using a certain set of features creates insensitivity towards other features. Only if the selected features are still present after a distribution shift, a model generalises o.o.d.

\section{Shortcut learning across deep learning}
\label{sec:across_deep_learning}
Taken together, we have seen how shortcuts are based on dataset shortcut opportunities and discriminative feature learing that result in a failure to generalise as intended. We will now turn to specific application areas, and discover how this general pattern appears across Computer Vision, Natural Language Processing, Agent-based (Reinforcement) Learning and Fairness / algorithmic decision-making. While shortcut learning is certainly not limited to these areas, they might be the most prominent ones where the problem has been observed.

\paragraph{Computer Vision}
To humans, for example, a photograph of a car still shows the same car even when the image is slightly transformed. To DNNs, in contrast, innocuous transformations can completely change predictions. This has been reported in various cases such as shifting the image by a few pixels \cite{azulay2018deep}, rotating the object \cite{alcorn2019strike}, adding a bit of random noise or blur \cite{Geirhos2018generalisation, dodge2019human, hendrycks2019benchmarking, michaelis2019benchmarking} or (as discussed earlier) by changing background \cite{beery2018recognition} or texture while keeping the shape intact \cite{geirhos2019imagenettrained} (see Figure~\ref{fig:generalisation} for examples).
Some key problems in Computer Vision are linked to shortcut learning. For example, transferring model performance across datasets (\emph{domain transfer}) is challenging because models often use domain-specific shortcut features, and shortcuts limit the usefulness of unsupervised representations \cite{minderer2020automatic}. Furthermore, \emph{adversarial examples} are particularly tiny changes to an input image that completely derail model predictions \cite{szegedy2013intriguing} (an example is shown in Figure~\ref{fig:generalisation}). Invisible to the human eye, those changes modify highly predictive patterns that DNNs use to classify objects \cite{ilyas2019adversarial}. In this sense, adversarial examples---one of the most severe failure cases of neural networks---can at least partly be interpreted as a consequence of shortcut learning.

\paragraph{Natural Language Processing}
The widely used language model BERT has been found to rely on superficial cue words. For instance, it learned that within a dataset of natural language arguments, detecting the presence of ``not'' was sufficient to perform above chance in finding the correct line of argumentation. This strategy turned out to be very useful for drawing a conclusion without understanding the content of a sentence \cite{niven2019probing}. Natural Language Processing suffers from very similar problems as Computer Vision and other fields. Shortcut learning starts from various dataset biases such as annotation artefacts \cite{goyal2017making, gururangan2018annotation, kaushik2018much, geva2019we}. Feature combination crucially depends on shortcut features like word length \cite{poliak2018hypothesis, kavumba2019choosing, niven2019probing, mccoy2019right}, and consequently leads to a severe lack of robustness such as an inability to generalise to more challenging test conditions \cite{agrawal2016analyzing, belinkov2017synthetic, jia2017adversarial, glockner2018breaking}. Attempts like incorporating a certain degree of unsupervised training as employed in prominent language models like BERT \cite{devlin2018bert} and GPT-2 \cite{radford2019gpt2} did not resolve the problem of shortcut learning \cite{niven2019probing}.

\paragraph{Agent-based (Reinforcement) Learning}
Instead of learning how to play Tetris, an algorithm simply learned to pause the game to evade losing \cite{vii2013first}. Systems of Agent-based Learning are usually trained using Reinforcement Learning and related approaches such as evolutionary algorithms. In both cases, designing a good reward function is crucial, since a reward function measures how close a system is to solving the problem. However, they all too often contain unexpected shortcuts that allow for so-called \emph{reward hacking} \cite{amodei2016concrete}. The existence of loopholes exploited by machines that follow the letter (and not the spirit) of the reward function highlight how difficult it is to design a shortcut-free reward function \cite{lehman2018surprising}. Reinforcement Learning is also a widely used method in Robotics, where there is a commonly observed \emph{generalisation} or \emph{reality gap} between simulated training environment and real-world use case. This can be thought of as a consequence of narrow shortcut learning by adapting to specific details of the simulation. Introducing additional variation in colour, size, texture, lighting, etc. helps a lot in closing this gap \cite{tobin2017domain, akkaya2019rubikscube}.

\paragraph{Fairness \& algorithmic decision-making}
Tasked to predict strong candidates on the basis of their résumés, a hiring tool developed by Amazon was found to be biased towards preferring men. The model, trained on previous human decisions, found gender to be such a strong predictor that even removing applicant names would not help: The model always found a way around, for instance by inferring gender from all-woman college names \cite{dastin2018amazon}. This exemplifies how some---but not all---problems of (un)fair algorithmic decision-making are linked to shortcut learning: Once a predictive feature is found by a model, even if it is just an artifact of the dataset, the model's decision rule may depend entirely on the shortcut feature. When human biases are not only replicated, but worsened by a machine, this is referred to as \emph{bias amplification} \cite{zhao2017men}. Other shortcut strategies include focusing on the majority group in a dataset while accepting high error rates for underrepresented groups \cite{buolamwini2018gender, rich2019lessons}, which can amplify existing societal disparities and even create new ones over time \cite{hashimoto2018fairness}. In the dynamical setting a related problem is called \emph{disparity amplification} \cite{hashimoto2018fairness}, where sequential feedback loops may amplify a model's reliance on a majority group. It should be emphasised, however, that fairness is an active research area of machine learning closely related to invariance learning that might be useful to quantify and overcome biases of both machine and human decision making.

\section{Diagnosing and understanding shortcut learning}
\label{sec:diagnosing_understanding}
Shortcut learning currently occurs across deep learning, causing machines to fail unexpectedly. Many individual elements of shortcut learning have been identified long ago by parts of the machine learning community and some have already seen substantial progress, but currently a variety of approaches are explored without a commonly accepted strategy. We here outline three actionable steps towards diagnosing and analysing shortcut learning.

\subsection{Interpreting results carefully}
\paragraph{Distinguishing datasets and underlying abilities} Shortcut learning is most deceptive when gone unnoticed. The most popular benchmarks in machine learning still rely on i.i.d. testing which drags attention away from the need to verify how closely this test performance measures the \emph{underlying ability} one is actually interested in. For example, the ImageNet dataset \cite{Russakovsky2015} was intended to measure the ability ``object recognition'', but DNNs seem to rely mostly on ``counting texture patches'' \cite{Brendel2019approximating}. Likewise, instead of performing ``natural language inference'', some language models perform well on datasets by simply detecting correlated key words \cite{gururangan2018annotation}. Whenever there is a discrepancy between the simplicity with which a dataset (e.g.\ ImageNet, SQuAD) can be solved and the complexity evoked by the high-level description of the underlying ability (e.g.\ object recognition, scene understanding, argument comprehension), it is important to bear in mind that a dataset is useful only for as long as it is a good proxy for the ability one is actually interested in \cite{gururangan2018annotation, zellers2019hellaswag}. We would hardly be intrigued by reproducing human-defined labels on datasets per se (a lookup table would do just as well in this case)---it is the underlying generalisation ability that we truly intend to measure, and ultimately improve upon.

\paragraph{Morgan's Canon for machine learning} Recall the cautionary tale of rats sniffing rather than seeing colour, described in Section~\ref{subsec:comparative_psychology}. Animals often trick experimenters by solving an experimental paradigm (i.e., dataset) in an unintended way without using the underlying ability one is actually interested in. This highlights how incredibly difficult it can be for humans to imagine solving a tough challenge in any other way than the human way: Surely, at Marr's implementational level \cite{marr1982vision} there may be differences between rat and human colour discrimination. But at the algorithmic level there is often a tacit assumption that human-like performance implies human-like strategy (or algorithm) \cite{borowski2019the}. This \emph{same strategy assumption} is paralleled by deep learning: Surely, DNN units are different from biological neurons---but if DNNs successfully recognise objects, it seems natural to assume that they are using object shape like humans do \cite{baker2018deep, Brendel2019approximating, geirhos2019imagenettrained}.

Comparative Psychology with its long history of comparing mental abilities across species has coined a term for the fallacy to confuse human-centered interpretations of an observed behaviour and the actual behaviour at hand (which often has a much simpler explanation): \emph{anthropomorphism}, ``the tendency of humans to attribute human-like psychological characteristics to nonhumans on the basis of insufficient empirical evidence'' \cite[p. 5]{buckner2019comparative}. As a reaction to the widespread occurrence of this fallacy, psychologist Lloyd Morgan developed a conservative guideline for interpreting non-human behaviour as early as 1903. It later became known as Morgan's Canon: ``In no case is an animal activity to be interpreted in terms of higher psychological processes if it can be fairly interpreted in terms of processes which stand lower on the scale of psychological evolution and development'' \cite[p. 59]{morgan1903introduction}. Picking up on a simple correlation, for example, would be considered a process that stands low on this psychological scale whereas ``understanding a scene'' would be considered much higher. It has been argued that Morgan's Canon can and should be applied to interpreting machine learning results \cite{buckner2019comparative}, and we consider it to be especially relevant in the context of shortcut learning. Accordingly, it might be worth acquiring the habit to confront machine learning models with a  ``Morgan's Canon for machine learning''\footnote{Our formulation is adapted from Hanlon's razor, ``Never attribute to malice that which can be adequately explained by stupidity''.}: \emph{Never attribute to high-level abilities that which can be adequately explained by shortcut learning.}

\paragraph{Testing (surprisingly) strong baselines}
In order to find out whether a result may also be explained by shortcut learning, it can be helpful to test whether a baseline model exceeds expectations even though it does not use intended features. Examples include using nearest neighbours for scene completion and estimating geolocation \cite{hays2007scene, hays2008im2gps}, object recognition with local features only \cite{Brendel2019approximating}, reasoning based on single cue words \cite{poliak2018hypothesis, niven2019probing} or answering questions about a movie without ever showing the movie to a model \cite{jasani2019are}. Importantly, this is not meant to imply that DNNs cannot acquire high-level abilities. They certainly do have the potential to solve complex challenges and serve as scientific models for prediction, explanation and exploration \cite{cichy2019deep}---however, we must not confuse performance on a \emph{dataset} with the acquisition of an \emph{underlying ability}.

\subsection{Detecting shortcuts: towards o.o.d.\ generalisation tests}
\paragraph{Making o.o.d.\ generalisation tests a standard practice}
Currently, measuring model performance by assessing validation performance on an i.i.d.\ test set is at the very heart of the vast majority of machine learning benchmarks. Unfortunately, in real-world settings the i.i.d.\ assumption is rarely justified; in fact, this assumption has been called ``the big lie in machine learning'' \cite{ghahramani2017panel}. While any metric is typically only an approximation of what we truly intend to measure, the i.i.d.\ performance metric may not be a good approximation as it can often be misleading, giving a false sense of security. In Section~\ref{subsec:education} we described how Bob gets a good grade on a multiple-choice exam through rote learning. Bob's reproductive approach gives the superficial appearance of excellent performance, but it would not generalise to a more challenging test. Worse yet, as long as Bob continues to receive good grades through surface learning, he is unlikely to change his learning strategy.

Within the field of Education, what is the best practice to avoid surface learning? It has been argued that changing the type of examination from multiple-choice tests to essay questions discourages surface learning, and indeed surface approaches typically fail on these kinds of exams \cite{scouller1998influence}. Essay questions, on the other hand, encourage so-called \emph{deep} or \emph{transformational} learning strategies \cite{marton1976qualitative, biggs1979individual}, like Alice's focus on understanding. This in turn enables transferring the learned content to \emph{novel} problems and consequently achieves a much better overlap between the educational objectives of the teacher and what the students actually learn \cite{chin2000learning}.
We can easily see the connection to machine learning---transferring knowledge to novel problems corresponds to testing generalisation beyond the narrowly learned setting \cite{marcus1998rethinking, kilbertus2018generalization, marcus2018deep}. If model performance is assessed only on i.i.d.\ test data, then we are unable to discover whether the model is actually acquiring the ability we think it is, since exploiting shortcuts often leads to deceptively good results on standard metrics \cite{lapuschkin2019unmasking}. We, among many others \cite{lake2016building, borowski2019the, chollet2019measure, crosby2019animal, juliani2019obstacle}, have explored a variety of o.o.d.\ tests and we hope it will be possible to identify a sufficiently simple and effective test procedure that could replace i.i.d. testing as a new standard method for benchmarking machine learning models in the future.

\definecolor{box.colbacktitle}{RGB}{156, 174, 211}
\definecolor{box.colback}{RGB}{217, 222, 238}
\begin{figure}[ht]
\small
\begin{tcolorbox}[width=\linewidth,colback={box.colback},
title={\textbf{Box I. EXAMPLES OF INTERESTING O.O.D.\ BENCHMARKS}},
colbacktitle=box.colbacktitle,coltitle=black]

We here list a few selected, encouraging examples of o.o.d.\ benchmarks.\\

\textbf{Adversarial attacks} can be seen as testing on model-specific worst-case o.o.d.\ data, which makes it an interesting diagnostic tool. If a successful adversarial attack \cite{szegedy2013intriguing} can change model predictions without changing semantic content, this is an indication that something akin to shortcut learning may be occurring \cite{ilyas2019adversarial, engstrom2019a}.\\

\textbf{ARCT with removed shortcuts} is a language argument comprehension dataset that follows the idea of removing known shortcut opportunities from the data itself in order to create harder test cases \cite{niven2019probing}.\\

\textbf{Cue conflict stimuli} like images with conflicting texture and shape information pitch features/cues against each other, such as an intended against an unintended cue  \cite{geirhos2019imagenettrained}. This approach can easily be compared to human responses.\\

\textbf{ImageNet-A} is a collection of natural images that several state-of-the-art models consistently classify wrongly. It thus benchmarks models on worst-case natural images \cite{hendrycks2019natural}.\\

\textbf{ImageNet-C} applies 15 different image corruptions to standard test images, an approach we find appealing for its variety and usability \cite{hendrycks2019benchmarking}.\\

\textbf{ObjectNet} introduces the idea of scientific controls into o.o.d.\ benchmarking, allowing to disentangle the influence of background, rotation and viewpoint \cite{barbu2019objectnet}.\\

\textbf{PACS} and other domain generalisation datasets require extrapolation beyond i.i.d.\ data per design by testing on a domain different from training data (e.g.\ cartoon images) \cite{li2017deeper}.\\

\textbf{Shift-MNIST / biased CelebA / unfair dSprites} are controlled toy datasets that introduce correlations in the training data (e.g.\ class-predictive pixels or image quality) and record the accuracy drop on clean test data as a way of finding out how prone a given architecture and loss function are to picking up on shortcuts \cite{heinze2017conditional, malhotra2018difference, creager2019flexibly, jacobsen2019excessive}.
\end{tcolorbox}
\end{figure}

\paragraph{Designing good o.o.d.\ tests}
While a distribution shift (between i.i.d.\ and o.o.d.\ data) has a clear mathematical definition, it can be hard to detect in practice \cite{recht2018cifar, recht2019imagenet}. In these cases, training a classifier to distinguish samples in dataset A from samples in dataset A' can reveal a distribution shift. We believe that good o.o.d.\ tests should fullfill at least the following three conditions: First, per definition there needs to be a \emph{clear distribution shift}, a shift that may or may not be distinguishable by humans. Second, it should have a \emph{well-defined intended solution}. Training on natural images while testing on white noise would technically constitute an o.o.d.\ test but lacks a solution. Third, a good o.o.d.\ test is a test where the majority of \emph{current models struggle}. Typically, the space of all conceivable o.o.d.\ tests includes numerous uninteresting tests. Thus given limited time and resources, one might want to focus on challenging test cases. As models evolve, generalisation benchmarks will need to evolve as well, which is exemplified by the Winograd Schema Challenge \cite{levesque2012winograd}. Initially designed to overcome shortcut opportunities caused by the open-ended nature of the Turing test, this common-sense reasoning benchmark was scrutinised after modern language models started to perform suspiciously well---and it indeed contained more shortcut opportunities than originally envisioned \cite{trichelair2019reasonable}, highlighting the need for revised tests. Fortunately, stronger generalisation tests are beginning to gain traction across deep learning. While o.o.d.\ tests will likely need to evolve alongside the models they aim to evaluate, a few current encouraging examples are listed in Box~I. In summary, rigorous generalisation benchmarks are crucial when distinguishing between the intended and a shortcut solution, and it would be extremely useful if a strong generally applicable testing procedure will emerge from this range of approaches.

\definecolor{solution.box.colbacktitle}{RGB}{190, 210, 165}
\definecolor{solution.box.colback}{RGB}{229, 237, 219}
\begin{figure}[h!]
\small
\begin{tcolorbox}[width=\linewidth,colback={solution.box.colback},title={\textbf{Box II. SHORTCUT LEARNING \& INDUCTIVE BIASES}},colbacktitle=solution.box.colbacktitle,coltitle=black]

The four components listed below determine the \emph{inductive bias} of a model and dataset: the set of assumptions that influence which solutions are learnable, and how readily they can be learned. Although in theory DNNs can approximate any function (given potentially infinite capacity) \cite{hornik1989multilayer}, their inductive bias plays an important role for the types of patterns they prefer to learn given finite capacity and data. 

\begin{itemize}
    \item \textbf{Structure: architecture.}
    Convolutions make it harder for a model to use location---a prior \cite{d2019finding} that is so powerful for natural images that even untrained networks can be used for tasks like image inpainting and denoising \cite{ulyanov2018deep}. In Natural Language Processing, transformer architectures \cite{vaswani2017attention} use \emph{attention layers} to understand the context by modelling relationships between words. In most cases, however, it is hard to understand the implicit priors in a DNN and even standard elements like ReLU activations can lead to unexpected effects like unwarranted confidence \cite{hein2019relu}.
    
    \item \textbf{Experience: training data.}
    As discussed in Section~\ref{subsec:dataset_shortcut_opportunities}, shortcut opportunities are present in most data and rarely disappear by adding more data \cite{jo2017measuring, lehman2018surprising, gururangan2018annotation, geirhos2019imagenettrained, ilyas2019adversarial}. Modifying the training data to block specific shortcuts has been demonstrated to work for reducing adversarial vulnerability \cite{madry2018towards} and texture bias \cite{geirhos2019imagenettrained}.

    \item \textbf{Goal: loss function.}
    The most commonly used loss function for classification, \emph{cross-entropy}, encourages DNNs to stop learning once a simple predictor is found; a modification can force neural networks to use all available information \cite{jacobsen2019excessive}. Regularisation terms that use additional information about the training data have been used to disentangle intended features from shortcut features \cite{heinze2017conditional, arjovsky2019invariant}. 

    \item \textbf{Learning: optimisation.}
    Stochastic gradient descent and its variants bias DNNs towards learning simple functions \cite{wu2017towards, de2018deep, valle-perez2018deep, sun2019lightlike}. The learning rate influences which patterns networks focus on: Large learning rates lead to learning simple patterns that are shared across examples, while small learning rates facilitate complex pattern learning and memorisation \cite{arpit2017closer, li2019towards}. The complex interactions between training method and architecture are poorly understood so far; strong claims can only be made for simple cases \cite{bartlett2019benign}.
\end{itemize}
\end{tcolorbox}
\end{figure}

\subsection{Shortcuts: why are they learned?}
\label{subsec:shortcuts_why_learned}

\paragraph{The ``Principle of Least Effort''}
Why are machines so prone to learning shortcuts, detecting grass instead of cows \cite{beery2018recognition} or a metal token instead of pneumonia \cite{zech2018variable}? Exploiting those shortcuts seems much \emph{easier} for DNNs than learning the intended solution. But what determines whether a solution is easy to learn? In Linguistics, a related phenomenon is called the ``Principle of Least Effort'' \cite{zipf1949human}, the observation that language speakers generally try to minimise the amount of effort involved in communication. For example, the use of ``plane'' is becoming more common than ``airplane'', and in pronouncing ``cupboard'', ``p'' and ``b'' are merged into a single sound \cite{ohala1990phonetics, vicentini2003economy}. Interestingly, whether a language change makes it easier for the speaker doesn't always simply depend on objective measures like word length. On the contrary, this process is shaped by a variety of different factors, including the anatomy (architecture) of the human speech organs and previous language experience (training data).

\paragraph{Understanding the influence of inductive biases}
In a similar vein, whether a solution is easy to learn for machines does not simply depend on the data but on all of the four components of a machine learning algorithm: architecture, training data, loss function, and optimisation. Often, the training process starts with feeding training data to the model with a fixed architecture and randomly initialised parameters. When the model's prediction is compared to ground truth, the loss function measures the prediction's quality. This supervision signal is used by an optimiser for adapting the model's internal parameters such that the model makes a better prediction the next time. Taken together, these four components (which determine the \emph{inductive bias} of a model) influence how certain solutions are much easier to learn than others, and thus ultimately determine whether a shortcut is learned instead of the intended solution \cite{sinz2019engineering}. Box II provides an overview of the connections between shortcut learning and inductive biases.

\section{Beyond shortcut learning}
\label{sec:beyond_shortcut_learning}

A lack of out-of-distribution generalisation can be observed all across machine learning. Consequently, a significant fraction of machine learning research is concerned with overcoming shortcut learning, albeit not necessarily as a concerted effort. Here we highlight connections between different research areas. Note that an exhaustive list would be out of the scope for this work. Instead, we cover a diverse set of approaches we find promising, each providing a unique perspective on learning beyond shortcut learning. \\

\noindent\textbf{Domain-specific prior knowledge} Avoiding reliance on unintended cues can be achieved by designing architectures and data-augmentation strategies that discourage learning shortcut features. If the orientation of an object does not matter for its category, either data-augmentation or hard-coded rotation invariance \cite{cohen2016group} can be applied. This strategy can be applied to almost any well-understood transformation of the inputs and finds its probably most general form in auto-augment as an augmentation strategy \cite{cubuk2019autoaugment}. Extreme data-augmentation strategies are also the core ingredient of the most successful semi-supervised \cite{berthelot2019mixmatch} and self-supervised learning approaches to date \cite{hjelm2018learning, oord2018representation}.\\

\noindent\textbf{Adversarial examples and robustness} Adversarial attacks are a powerful analysis tool for worst-case generalisation \cite{szegedy2013intriguing}. Adversarial examples can be understood as counterfactual explanations, since they are the smallest change to an input that produces a certain output. Achieving counterfactual explanations of predictions aligned with human intention makes the ultimate goals of adversarial robustness tightly coupled to causality research in machine learning \cite{scholkopf2019causality}. Adversarially robust models are somewhat more aligned with humans and show promising generalisation abilities \cite{schott2018towards, engstrom2019learning}. While adversarial attacks test model performance on model-dependent worst-case noise, a related line of research focuses on model-independent noise like image corruptions \cite{Geirhos2018generalisation, hendrycks2019benchmarking}.\\

\noindent\textbf{Domain adaptation, -generalisation and -randomisation} These areas are explicitly concerned with out-of-distribution generalisation. Usually, multiple distributions are observed during training time and the model is supposed to generalise to a new distribution at test time. Under certain assumptions the intended (or even causal) solution can be learned from multiple domains and environments \cite{peters2016causal, heinze2017conditional, arjovsky2019invariant}. In robotics, domain randomisation (setting certain simulation parameters randomly during training) is a very successful approach for learning policies that generalise to similar situations in the real-world \cite{tobin2017domain}.\\

\noindent\textbf{Fairness} Fairness research aims at making machine decisions ``fair'' according to a certain definition \cite{dwork2012fairness}. Individual fairness aims at treating similar individuals similarly while group fairness aims at treating subgroups no different than the rest of the population \cite{zemel2013learning, hardt2016equality}. Fairness is closely linked to generalisation and causality \cite{kusner2017counterfactual}. Sensitive group membership can be viewed as a domain indicator: Just like machine decisions should not typically be influenced by changing the domain of the data, they also should not be biased against minority groups.\\

\noindent\textbf{Meta-learning} Meta-learning seeks to learn how to learn. An intermediate goal is to learn representations that can adapt quickly to new conditions \cite{schmidhuber1987evolutionary, santoro2016meta, finn2017model}. This ability is connected to the identification of causal graphs \cite{bengio2019meta} since learning causal features allows for small changes when changing environments.\\

\noindent\textbf{Generative modelling and disentanglement} Learning to generate the observed data forces a neural network to model every variation in the training data. By itself, however, this does not necessarily lead to representations useful for downstream tasks \cite{Fetaya2020Understanding}, let alone out-of-distribution generalisation. Research on disentanglement addresses this shortcoming by learning generative models with well-structured latent representations \cite{higgins2017beta}. The goal is to recover the true generating factors of the data distribution from observations \cite{hyvarinen2000independent} by identifying independent causal mechanisms \cite{scholkopf2019causality}.

\section{Conclusion}
\begin{flushright}
\emph{``The road reaches every place, the short cut only one''}\\
--- James Richardson \cite{richardson2001vectors}
\end{flushright}

\noindent
Science aims for understanding. While deep learning as an engineering discipline has seen tremendous progress over the last few years, deep learning as a scientific discipline is still lagging behind in terms of understanding the principles and limitations that govern how machines learn to extract patterns from data. A deeper understanding of how to overcome shortcut learning is of relevance beyond the current application domains of machine learning and there might be interesting future opportunities for cross-fertilisation with other disciplines such as Economics (designing management incentives that do not jeopardise long-term success by rewarding unintended ``shortcut'' behaviour) or Law (creating laws without ``loophole'' shortcut opportunities). Until the problem is solved, however, we offer the following four recommendations:

\paragraphwithlinebreak{(1) Connecting the dots: shortcut learning is ubiquitous}
Shortcut learning appears to be a ubiquitous characteristic of learning systems, biological and artificial alike. Many of deep learning's problems are connected through shortcut learning---models exploit dataset shortcut opportunities, select only a few predictive features instead of taking all evidence into account, and consequently suffer from unexpected generalisation failures. ``Connecting the dots'' between affected areas is likely to facilitate progress, and making progress can generate highly valuable impact across various applications domains. 

\paragraphwithlinebreak{(2) Interpreting results carefully}
Discovering a shortcut often reveals the existence of an easy solution to a seemingly complex dataset. We argue that we will need to exercise great care before attributing high-level abilities like ``object recognition'' or ``language understanding'' to machines, since there is often a much simpler explanation.

\paragraphwithlinebreak{(3) Testing o.o.d.\ generalisation}
Assessing model performance on i.i.d.\ test data (as the majority of current benchmarks do) is insufficient to distinguish between intended and unintended (shortcut) solutions. Consequently, o.o.d.\ generalisation tests will need to become the rule rather than the exception.

\paragraphwithlinebreak{(4) Understanding what makes a solution easy to learn}
DNNs always learn the easiest possible solution to a problem, but understanding which solutions are easy (and thus likely to be learned) requires disentangling the influence of structure (architecture), experience (training data), goal (loss function) and learning (optimisation), as well as a thorough understanding of the interactions between these factors.\\

\noindent
Shortcut learning is one of the key roadblocks towards fair, robust, deployable and trustworthy machine learning. While overcoming shortcut learning in its entirety may potentially be impossible, any progress towards mitigating it will lead to a better alignment between learned and intended solutions. This holds the promise that machines behave much more reliably in our complex and ever-changing world, even in situations far away from their training experience. Furthermore, machine decisions would become more transparent, enabling us to detect and remove biases more easily. Currently, the research on shortcut learning is still fragmented into various communities. With this perspective we hope to fuel discussions across these different communities and to initiate a movement that pushes for a new standard paradigm of generalisation that is able to replace the current i.i.d.\ tests.

\subsubsection*{Acknowledgement}
\begin{footnotesize}
The authors thank the International Max Planck Research School for Intelligent Systems (IMPRS-IS) for supporting R.G. and C.M.; the Deutsche Forschungsgemeinschaft (DFG, German Research Foundation) for supporting C.M. via grant EC 479/1-1; the Collaborative Research Center (Projektnummer 276693517---SFB 1233: Robust Vision) for supporting M.B. and F.A.W.; the German Federal Ministry of Education and Research through the Tübingen AI Center (FKZ 01IS18039A) for supporting W.B. and M.B.; as well as the Natural Sciences and Engineering Research Council of Canada and the Intelligence Advanced Research Projects Activity (IARPA) via Department of Interior/Interior Business Center (DoI/IBC) contract number D16PC00003 for supporting J.J.

The authors would like to thank Judy Borowski, Max Burg, Santiago Cadena, Alexander S. Ecker, Lisa Eisenberg, Roland Fleming, Ingo Fr\"und, Samuel Greiner, Florian Grießer, Shaiyan Keshvari, Ruth Kessler, David Klindt, Matthias K\"ummerer, Benjamin Mitzkus, Hendrikje Nienborg, Jonas Rauber, Evgenia Rusak, Steffen Schneider, Lukas Schott, Tino Sering, Yash Sharma, Matthias Tangemann, Roland Zimmermann and Tom Wallis for helpful discussions. %
\end{footnotesize}

\subsubsection*{Author contributions}
\begin{footnotesize}
The project was initiated by R.G. and C.M. and led by R.G. with support from C.M. and J.J.; M.B. and W.B. reshaped the initial thrust of the perspective and together with R.Z. supervised the machine learning components. The toy experiment was conducted by J.J. with input from R.G. and C.M. Most figures were designed by R.G. and W.B. with input from all other authors. Figure 2 (left) was conceived by M.B. The first draft was written by R.G., J.J. and C.M. with input from F.A.W. All authors contributed to the final version and provided critical revisions from different perspectives.
\end{footnotesize}

\bibliographystyle{naturemag}
\bibliography{refs}

\newpage
\renewcommand{\thesubsection}{\Alph{subsection}}
\section*{Appendix}
\appendix

\subsection{Toy example: method details}
\label{appendix:toy_example}
\begin{footnotesize}
The code to reproduce our toy example (Figure~\ref{fig:toy_example}) is available from \url{https://github.com/rgeirhos/shortcut-perspective}. Two easily distinguishable shapes (star and moon) were placed on a $200 \times 200$ dimensional 2D canvas. The training set is constructed out of 4000 images, where 2000 contain a star shape and 2000 a moon shape. The star shape is randomly placed in the top right and bottom left quarters of the canvas, whereas the moon shape is randomly placed in the top left and bottom right quarters of the canvas. At test time the setup is nearly identical, 1000 images with a star and 1000 images with a moon are presented. However, this time the position of star and moon shapes are randomised over the full canvas, i.e.\ in test images stars and moons can appear at any location.

We train two classifiers on this dataset: a fully connected network as well as a convolutional network. The classifiers are trained for five epochs with a batch size of 100 on the training set and evaluated on the test set. The training objective is standard crossentropy loss and the optimizer is Adam with a learning rate of 0.00001, $\beta_1=0.9$, $\beta_2= 0.999$ and $\epsilon=1e-08$. The fully connected network was a three-layer ReLU MLP (multilayer perceptron) with 1024 units in each layer and two output units corresponding to the two target classes. It reaches 100\% accuracy at training time and approximately chance-level accuracy at test time (51.0\%). The convolutional network had three convolutional layers with 128 channels, a stride of 2 and filter size of $5 \times 5$ interleaved with ReLU nonlinearities, followed by a global average pooling and a linear layer mapping the 128 outputs to the logits. It reaches 100\% accuracy on train and test set.
\end{footnotesize}

\subsection{Image rights \& attribution}
\begin{footnotesize}
Figure~\ref{fig:teaser} consists of four images from different sources. The first image from the left was taken from \url{https://aiweirdness.com/post/171451900302/do-neural-nets-dream-of-electric-sheep} with permission of the author. The second image from the left was generated by ourselves. The third image from the left is from  ref.~\cite{zech2018variable}. It was released under the \href{https://creativecommons.org/licenses/by/4.0/}{CC BY 4.0 license} as stated here: \url{https://journals.plos.org/plosmedicine/article?id=10.1371/journal.pmed.1002683} and adapted by us from \href{https://doi.org/10.1371/journal.pmed.1002683.g002}{Figure 2B} of the corresponding publication. The image on the right is Figure 1 from ref.~\cite{jia2017adversarial}. It was released under \href{https://creativecommons.org/licenses/by/4.0/}{CC BY 4.0 license} as stated here: \url{https://www.aclweb.org/anthology/D17-1215/}(at the bottom) and retrieved by us from \href{https://www.aclweb.org/anthology/D17-1215.pdf}.

The image from Section~\ref{subsec:dataset_shortcut_opportunities} was adapted from Figure 1 of ref.~\cite{beery2018recognition} with permission from the authors (image cropped from original figure by us). The image from Section~\ref{subsec:feature_combination} was adapted from Figure 1 of ref.~\cite{geirhos2019imagenettrained} with permission from the authors (image cropped from original figure by us).
The image from Section~\ref{subsec:generalisation_robustness} was adapted from Figure 1 of ref.~\cite{nguyen2015deep} with permission from the authors (image cropped from original figure by us).

Figure~\ref{fig:generalisation} consists of a number of images from different sources. The first author of the corresponding publication is mentioned in the figure for identification. The images from ref.~\cite{szegedy2013intriguing} were released under the \href{https://creativecommons.org/licenses/by/3.0/legalcode}{CC BY 3.0 license} as stated here: \url{https://arxiv.org/abs/1312.6199} and adapted by us from Figure 5a of the corresponding publication (images cropped from original figure by us). The images from ref.~\cite{dodge2019human} were adapted from Figure 1 of the corresponding paper with permission from the authors (images cropped from original figure by us). The images from ref.~\cite{alcorn2019strike} were adapted from Figure 1 of the corresponding paper with permission from the authors (images cropped from original figure by us). The images from ref.~\cite{geirhos2019imagenettrained} were adapted from Figure 1 of the corresponding paper with permission from the authors (images cropped from original figure by us). The images from ref.~\cite{jacobsen2019excessive} were adapted from Figure 1 of the corresponding paper with permission from the authors (images cropped from original figure by us). The images from ref.~\cite{Brendel2019approximating} were adapted from Figure 5 of the corresponding paper with permission from the authors (images cropped from original figure by us). The images from ref.~\cite{beery2018recognition} were adapted from Figure 1 of the corresponding paper with permission from the authors (images cropped from original figure by us). The images from ref.~\cite{nguyen2015deep} were adapted from Figure 1 and Figure 2 of the corresponding paper with permission from the authors (images cropped from original figures by us).
\end{footnotesize}

\end{document}